\documentclass[letterpaper,journal]{IEEEtran}
\usepackage{cite}
\usepackage{amsmath,amssymb,amsfonts}
\usepackage{multirow}
\usepackage{booktabs}
\usepackage{array}
\usepackage{algorithm}
\usepackage{algorithmic}
\usepackage{graphicx}
\usepackage[caption=false,font=footnotesize]{subfig}
\usepackage{textcomp}
\usepackage{stfloats}
\usepackage{url}
\usepackage{microtype}
\usepackage{needspace}
\usepackage[hidelinks,hyperfootnotes=false]{hyperref}

\begin{document}

\title{Training-free Acceleration for Blind Compressed Image Restoration via Activation-Calibrated Singular Value Decomposition}

\author{Suhang Yao, Xin Li, Xijun Wang, Yufei Wang, Yeying Jin, and Zhibo Chen%
\thanks{\textit{(Corresponding authors: Xin Li and Zhibo Chen.)}}%
\thanks{Suhang Yao, Xin Li, Xijun Wang, and Zhibo Chen are with the University of Science and Technology of China, Hefei, China (e-mail: yshustc@mail.ustc.edu.cn; xin.li@ustc.edu.cn; xijunwang00@gmail.com; chenzhibo@ustc.edu.cn).}%
\thanks{Yufei Wang is with SparcAI Inc. (e-mail: yufei.wang@sparclab.ai).}%
\thanks{Yeying Jin is with the National University of Singapore, Singapore (e-mail: jinyeying@u.nus.edu).}%
}

\pagestyle{plain}

\maketitle
\thispagestyle{plain}

\begin{figure*}[b]
\normalfont\fontsize{8}{9.5}\selectfont
\noindent\copyright\ 2026 IEEE. Personal use of this material is permitted. Permission from IEEE must be obtained for all other uses, in any current or future media, including reprinting/republishing this material for advertising or promotional purposes, creating new collective works, for resale or redistribution to servers or lists, or reuse of any copyrighted component of this work in other works.\par
\noindent DOI: \url{https://doi.org/10.1109/JETCAS.2026.3745896}
\end{figure*}

\begin{abstract}
Transformer-based models achieve strong performance on blind compressed image restoration (CIR) under unknown compression settings, but their densely stacked projection operators make deployment costly.
We study training-free acceleration for such models through an empirical, operator-level analysis.
Specifically, we profile where computation is concentrated, characterize channel-wise activation distributions under compressed inputs, and examine the singular-value decay of representative projection weights.
These measurements identify the recurring projections in the attention, feed-forward, convolutional, and prompt-conditioning pathways as effective low-rank replacement targets that preserve the original inference graph.
They further establish truncated SVD (TSVD) as a principled training-free baseline for blind CIR and reveal why a uniform low-rank objective is mismatched to the channel-imbalanced feature distributions of compressed images.
Accordingly, we formulate Activation-Calibrated Singular Value Decomposition (ASVD) for this restoration setting, deriving low-rank factors \emph{entirely offline} from pretrained weights and lightweight activation statistics gathered from only a few compressed calibration images.
By weighting the decomposition with channel-wise activation magnitudes, our ASVD assigns larger penalties to errors on more active channels, requiring neither retraining nor online calibration and leaving the restoration pipeline intact.
When instantiated on PromptCIR at rank~64, ASVD consistently improves PSNR-B over TSVD across four standard benchmarks under identical parameter and MAC budgets.
At this operating point, the factorized model cuts parameters by $59\%$ ($34.75$M\,$\rightarrow$\,$14.34$M) and computation by $16.1\%$ ($258.79$\,$\rightarrow$\,$217.16$ GMACs for a $256\times256$ input), enabling acceleration without retraining.
\end{abstract}

\section{Introduction}
\label{sec:intro}

Lossy image compression is indispensable for large-scale visual storage and transmission.
JPEG remains one of the most widely deployed standards thanks to its simplicity and broad compatibility, yet aggressive compression inevitably introduces blocking, ringing, and texture loss.
Blind compressed image restoration (CIR) targets this problem by recovering faithful images from compressed inputs without access to the compression quality factor or other degradation parameters.

Deep restoration models have substantially advanced blind CIR.
Transformer-based architectures are particularly effective, since self-attention and hybrid attention blocks capture the long-range dependencies that purely local models struggle to represent~\cite{ali2023vision}.
Representative backbones such as Uformer~\cite{wang2022uformer}, Restormer~\cite{zamir2022restormer}, and HAT~\cite{chen2023hat} have performed well across a range of low-level vision tasks.
For blind CIR, PromptCIR~\cite{li2024promptcir} shows that prompt-based conditioning improves robustness when compression artifacts are unknown and spatially varying.

The architectural choices that raise restoration quality also complicate deployment.
Modern transformer-based CIR models repeatedly apply high-dimensional projection operators in their attention, feed-forward, convolutional-projection, and prompt-conditioning pathways.
Because these operators recur across many scales and layers, they dominate both parameters and computation.
Our profiling of PromptCIR shows that the attention, feed-forward, and convolutional-projection operators alone account for $99.43\%$ of parameters and $99.89\%$ of convolutional and linear MACs in the $128\times128$ profiling setting, which makes them the primary targets for acceleration.
For resource-constrained or latency-sensitive pipelines, reducing these projections directly improves efficiency while keeping the surrounding restoration process intact.

We approach acceleration from this operator-level evidence rather than from a fixed compression recipe.
An effective method should target the repeatedly used projections, preserve the pretrained restoration pipeline, avoid task-specific retraining, and account for channel-dependent behavior under compressed inputs.
We therefore examine activation statistics and singular-value spectra before deciding how to replace these operators.
The spectral analysis points to truncated singular value decomposition (TSVD) as a natural training-free baseline: linear projections and reshaped convolutional projections are written as matrices, decomposed offline, and replaced by two sequential low-rank projections.
Direct TSVD, however, treats all channels uniformly under an unweighted matrix objective.
This assumption is restrictive for restoration, since compressed-image features exhibit markedly nonuniform channel magnitudes tied to artifacts, textures, and prompt-conditioned information.

Guided by these observations, we design \textbf{Activation-Calibrated Singular Value Decomposition (ASVD)} as a projection-level acceleration method for transformer-based blind CIR.
The method first expresses the selected linear and convolutional projections in a unified matrix form.
TSVD then provides a closed-form, training-free low-rank baseline grounded in the Eckart--Young theorem~\cite{eckart1936lora}.
Building on this baseline, our ASVD uses lightweight input and output activation statistics collected from a small set of compressed calibration images to weight the decomposition.
The resulting low-rank factors are produced entirely offline and inserted back into the pretrained network, requiring no ground-truth restored images, no retraining, no prompt redesign, and no online calibration.

We instantiate the framework on PromptCIR, a representative prompt-based blind CIR model, and evaluate it on standard blind CIR benchmarks.

At rank~64, ASVD improves PSNR-B over TSVD across all four benchmarks while maintaining comparable PSNR and SSIM under identical parameter and MAC budgets.

The main contributions are summarized as follows:
\begin{itemize}
	\item We present a training-free, projection-level low-rank acceleration formulation for transformer-based blind CIR. Operator profiling and singular-value analysis identify the dominant replacement targets and establish TSVD as an architecture-preserving baseline.
	\item We design ASVD to address the channel imbalance observed in compressed-image feature activations. Our decomposition draws on lightweight statistics from compressed calibration inputs to emphasize activation-sensitive channels, without paired targets or retraining.
	\item We instantiate the method on PromptCIR and evaluate it against the original backbone and TSVD. The results show consistent PSNR-B improvements over TSVD at rank~64 under the same parameter and MAC budgets, while leaving the surrounding restoration pipeline unchanged.
\end{itemize}
\section{Related Work}
\label{sec:related}

\subsection{Compressed Image Restoration}

Compressed image restoration (CIR) targets artifacts caused by quantization and transform loss.
Early approaches relied on transform-domain priors or dual-domain modeling~\cite{liu2015data}, using handcrafted assumptions to recover information lost to compression.

CNN-based methods later pushed CIR performance further.
ARCNN~\cite{dong2015arcnn} demonstrated the effectiveness of convolutional networks for artifact reduction, while dual-domain designs such as D3~\cite{wang2016d3} and DMCNN~\cite{zhang2018dmcnn} exploited complementary representations in the pixel and transform domains, following the residual learning paradigm~\cite{he2016deep,zhang2017dncnn,mao2016red,tai2017memnet}.
Attention-enhanced CNN restorers further improved representation capacity through channel and non-local modeling~\cite{zhang2018rcan,zhang2019rnan}, while quantization-guided designs injected JPEG side information~\cite{ehrlich2020qgac}.

More recently, transformer-based architectures have become popular restoration backbones, owing to the global modeling capability of self-attention~\cite{vaswani2017attention}.
Representative models such as SwinIR~\cite{liang2021swinir}, Uformer~\cite{wang2022uformer}, Restormer~\cite{zamir2022restormer}, and HAT~\cite{chen2023hat,hat} handle long-range dependencies and complex artifact patterns effectively, while Stripformer~\cite{tsai2022stripformer} and NAFNet~\cite{chen2022simple} improve efficiency through structured attention or simplified operators.
State-space restoration backbones such as MambaIR~\cite{guo2024mambair} offer another route to efficient long-range modeling.
This progression has been evaluated across diverse restoration benchmarks and datasets~\cite{ali2023vision,li2023lsdir,DIV2K,ICB,yang2022aim}.
However, the gains of these backbones come largely from stacking high-dimensional projection and attention operators, whose cost grows rapidly with feature width and token length, making efficient deployment a recurring concern.

The blind CIR setting poses additional challenges, because the compression artifacts are unknown and spatially varying.
To improve robustness, prior work incorporates auxiliary cues or representation learning strategies, including quality-factor-aware modeling~\cite{jiang2021fbcnn} and contrastive representation learning~\cite{wang2022jpeg}.
More recently, prompt-based conditioning has emerged as an effective way to handle diverse degradations, with prompt-conditioned transformers proving effective in both all-in-one and blind CIR settings, including PromptIR~\cite{potlapalli2023promptir}, PromptCIR~\cite{li2024promptcir}, ProRes~\cite{ma2023prores}, PromptRR~\cite{wang2024promptrr}, Prompt-in-Prompt~\cite{pip}, DA-CLIP~\cite{daclip}, and visual or multimodal prompt learning methods~\cite{visual_prompt_tuning,prompt_learning,visual_prompt_tuning2}.
Related compressed-image restoration and super-resolution efforts, including UCIP~\cite{li2024ucip}, MambaCSR~\cite{ren2024mambacsr}, and CODiff~\cite{guo2025compression}, further show the importance of compression-aware conditioning.
Such prompt conditioning introduces additional projection pathways that interact with the backbone, further enlarging the operator budget that any acceleration method must address.

\subsection{Model Compression and Acceleration}

Model efficiency has been studied extensively through pruning, quantization, and low-rank approximation~\cite{liang2021pruning,han2015learning,han2015deep,liu2017learning}.
Among these, low-rank approximation offers a principled and training-free route to reducing computation, grounded in the classical Eckart--Young theorem~\cite{eckart1936lora}.
Classical CNN acceleration methods showed that convolutional redundancy can be exploited through low-rank or tensor factorization~\cite{denton2014exploiting,jaderberg2014speeding,lebedev2014speeding}.
Recent works revisit truncated SVD for large neural networks and propose truncation-aware~\cite{wang2024svd} or importance-weighted and Fisher-weighted variants~\cite{chekalina2025fwsvd} to improve approximation fidelity.
A shared insight is that raw weight-reconstruction error is an incomplete objective, since channels contribute unequally to the layer output; this motivates activation- or importance-aware reweighting of the decomposition objective.

In parallel, training-free acceleration techniques for iterative generative models exploit feature reuse and operator-level approximation.
Representative examples include DeepCache~\cite{ma2024deepcache}, block-wise caching~\cite{wimbauer2024blockcache}, and channel-aware SVD acceleration for diffusion transformers~\cite{chen2025icc}.
Other cache-based or training-free variants further explore layer caching, token-wise feature reuse, fast-forward caching, and delta updates~\cite{selvaraju2024fora,ma2024learning,huang2024harmonica,chen2024delta,zou2024accelerating}.
Although developed mainly for diffusion models, these approaches reinforce calibration and operator-level approximation as practical tools for training-free acceleration.

We develop a training-free acceleration scheme for PromptCIR-based blind CIR that calibrates projection factorization using activation magnitudes from compressed images. The resulting diagonal scaling provides a lightweight alternative to truncation-aware whitening~\cite{wang2024svd} and Fisher-weighted objectives~\cite{chekalina2025fwsvd}. Whereas channel-aware decomposition for diffusion transformers supports cache correction~\cite{chen2025icc}, our factors directly replace spatial and prompt-conditioning projections, preserving the restoration pipeline without retraining.
\section{Motivation and Design Rationale}
\label{sec:motivation}

We derive the projection-level SVD design from an empirical analysis of transformer-based blind CIR.
The analysis is organized around three questions: which operators dominate the cost, whether their weights admit a compact low-rank approximation, and whether compressed-image activations justify channel-aware weighting.
The answers determine our method design: projection operators define the replacement unit, TSVD supplies the training-free baseline, and ASVD supplies activation-aware weighting.

\begin{figure*}[t]
	\centering
	\includegraphics[width=\textwidth]{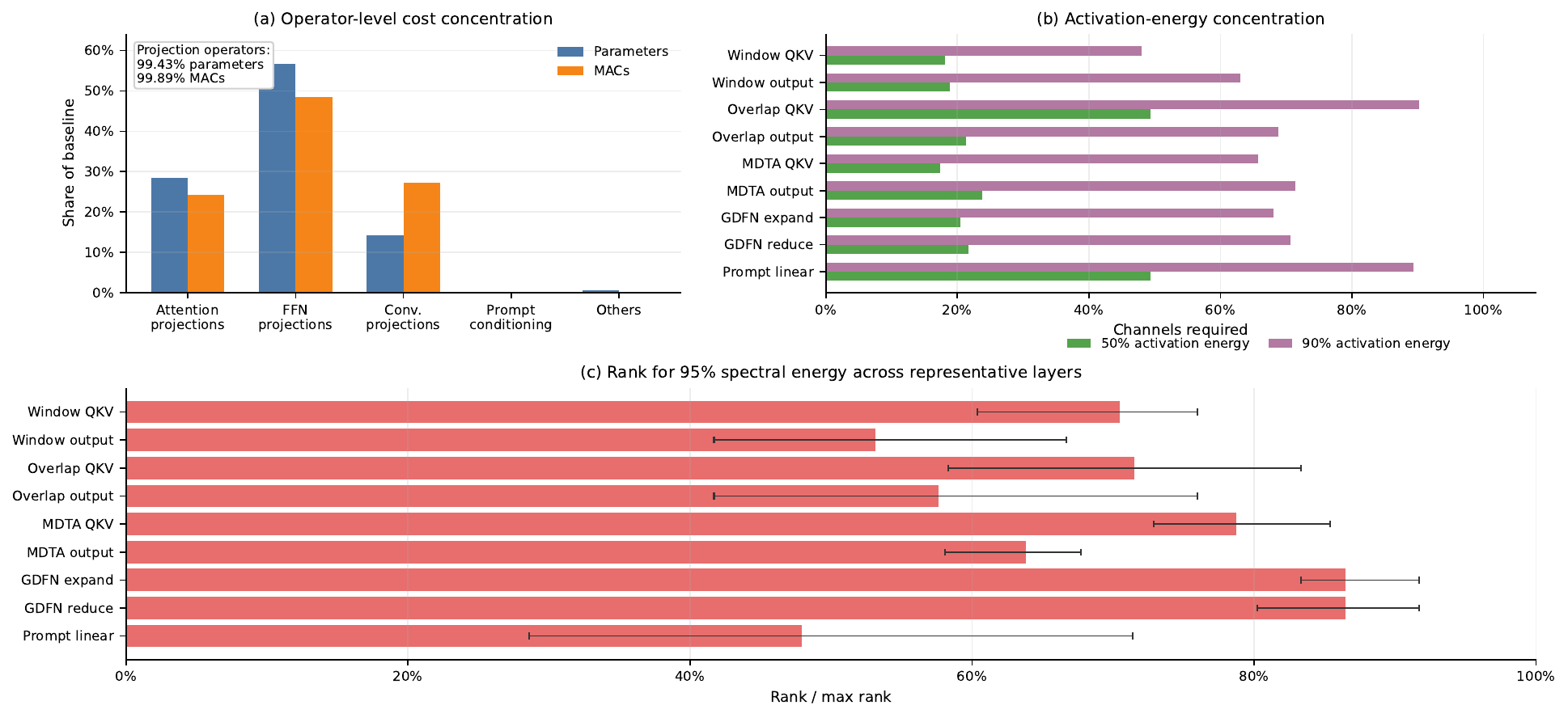}
	\caption{Empirical motivation for projection-level ASVD.
		(a) Parameter distribution and convolutional and linear MAC distribution across operator groups in PromptCIR, evaluated at an input resolution of \(128\times128\).
		(b) Output activations are channel-imbalanced: a relatively small fraction of channels often accounts for 50\% of activation energy, while the fraction required for 90\% energy varies by operator family.
		(c) The rank required to preserve 95\% spectral energy also varies across operators; bars show the mean over representative layers and error bars show the min--max range.}
	\label{fig:motivation_analysis}
\end{figure*}

\subsection{Projection Operators as the Replacement Unit}
\label{sec:projection_unit}

Prompt-based transformer restoration models contain many projection operators, including query-key-value projections, output projections, feed-forward expansions and reductions, point-wise convolutional projections, and prompt-conditioning projections.
These modules are invoked repeatedly across the encoder, bottleneck, and decoder stages.
They share a unified matrix representation: a linear layer is already a matrix, and a convolution kernel can be reshaped into a two-dimensional operator.
This common representation localizes the replacement while preserving the architecture.

The cost profile in Fig.~\ref{fig:motivation_analysis}(a) shows that this choice is not merely convenient algebraically but also meaningful for efficiency.
In the profiled PromptCIR model, attention projections, feed-forward projections, and convolutional-projection layers together account for $99.43\%$ of parameters and $99.89\%$ of convolutional and linear MACs at $128\times128$, so replacement can concentrate on these operators rather than on the non-dominant modules.
Prompt-conditioning operators are far smaller in aggregate, yet we retain them within the unified formulation because they govern degradation-conditioned feature modulation.
Replacing selected projections is also far less intrusive than redesigning attention blocks.
The surrounding normalization, nonlinearity, skip-connection, and prompt-conditioning logic can remain unchanged, while each dense projection is realized by two low-rank factors.
This locality is essential in an offline replacement setting, where the compressed model must reproduce the pretrained restoration mapping as closely as possible.

\subsection{Spectral Evidence for SVD-Based Replacement}
\label{sec:svd_rationale}

After identifying the replacement unit, we examine whether SVD is a reasonable low-rank mechanism for these operators.
SVD satisfies the deployment constraints imposed by the pretrained restoration model: it can be computed offline from pretrained weights, yields the optimal rank-$r$ approximation under the standard Frobenius objective, and applies uniformly to linear projections and reshaped convolution kernels.
It preserves the surrounding inference graph, since each original projection is replaced by two sequential low-rank projections with the same tensor interface.

The singular-value analysis in Fig.~\ref{fig:motivation_analysis}(c) shows why SVD is an appropriate starting point but not a complete design.
Representative projection operators exhibit measurable spectral concentration, but its degree varies across operator families: the mean rank needed to preserve $95\%$ of the spectral energy, defined as the sum of squared singular values, ranges from $47.93\%$ for prompt linear projections to $86.47\%$ for GDFN projections.
This evidence supports SVD-based replacement while indicating that rank must be analyzed explicitly.
First, TSVD is a principled baseline because it tests how much of the pretrained projection behavior a bounded-rank subspace can capture.
Second, because uniform low-rank replacement is not equally effective for all projections, rank and operator selection should be evaluated explicitly rather than fixed globally.

\subsection{Channel Statistics Under Compressed Inputs}
\label{sec:calibration_need}

TSVD provides the low-rank baseline, but its objective is uniform over channels.
For blind CIR, this assumption deserves scrutiny because the network must preserve spatially precise structures while suppressing compression artifacts.
As compressed images pass through a restoration transformer, different channels carry different amounts of artifact, texture, and prompt-conditioned information.
We therefore inspect the activation statistics of high-cost projections using a small compressed calibration set.

The observed activation distributions are markedly nonuniform.
For several window-attention, multi-Dconv head transposed attention (MDTA), gated-Dconv feed-forward network (GDFN), and output-projection families, $50\%$ of the output activation energy, defined as the sum of squared per-channel activation magnitudes, concentrates in only $17.39\%$--$23.76\%$ of the channels, whereas prompt-related and some overlap-attention projections are less concentrated.
Such heterogeneity means that a single unweighted objective treats all channels alike even when their activation magnitudes differ sharply under compressed inputs, which makes uniform channel treatment restrictive.
If a high-activation channel is poorly approximated, the resulting error can propagate through the restoration features far more strongly than an error on a weakly activated channel.

Our ASVD uses this observation as calibration information rather than as a training signal.
The collected statistics neither optimize a restoration loss, nor rely on ground-truth restored images, nor modify the model online.
Instead, they define diagonal scaling factors that give activation-dominant input and output channels greater influence within a deterministic offline replacement procedure.

\subsection{From Analysis to ASVD}
\label{sec:analysis_to_asvd}

These observations define a two-step replacement strategy.
Projection-level replacement targets the dominant cost centers while keeping the pretrained restoration pipeline intact.
The spectral results support TSVD as a closed-form low-rank baseline for these matrix-representable operators.
The activation results then motivate ASVD, our factor-generation refinement that introduces activation-derived channel weights so that channels more active under compressed inputs gain greater influence in the weighted approximation objective.
At a fixed rank, TSVD and ASVD share the same inference-time structure, parameter count, and MACs; comparing them therefore isolates the effect of activation-calibrated factor generation under an identical deployment budget.
\section{Methodology}
\label{sec:method}

\begin{figure*}[t]
	\centering
	\includegraphics[width=0.98\textwidth,trim=7bp 117bp 7bp 101bp,clip]{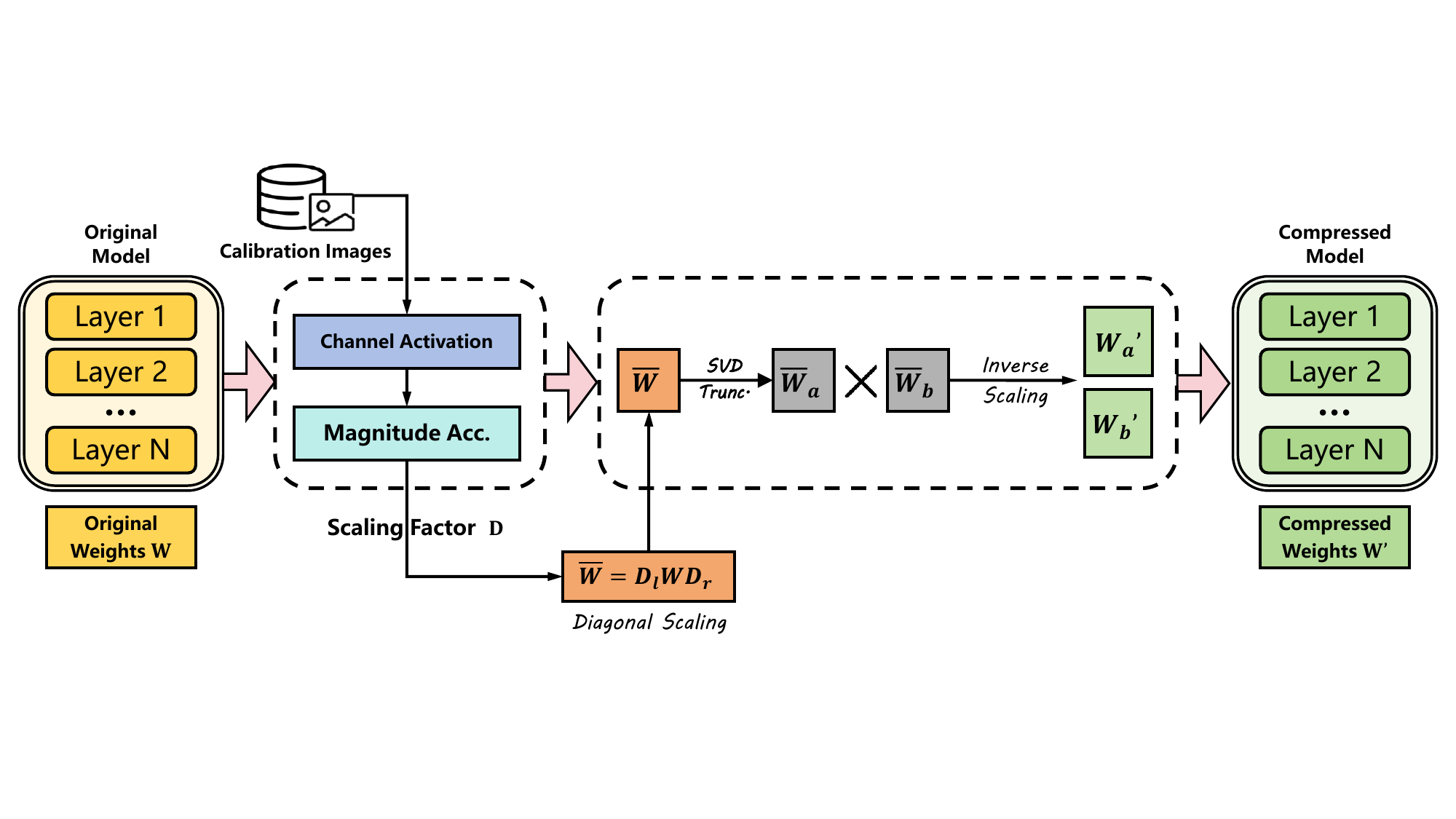}
\caption{Overview of our ASVD. We select high-cost projection operators in a transformer-based blind CIR model, collect lightweight per-channel activation statistics on a small calibration set, and form diagonal scaling factors. Truncated SVD is applied to the scaled weights, and the low-rank factors are mapped back via inverse scaling, yielding factorized replacements that preserve the original inference flow.}
	\label{fig:asvd_overview}
\end{figure*}

\subsection{Overview}
\label{sec:overview}

The analysis in Sec.~\ref{sec:motivation} leads to an operator-replacement formulation: selected projections are replaced by two low-rank factors generated offline from pretrained weights and, in our ASVD variant, lightweight activation statistics.
The surrounding restoration pipeline is left unchanged, and no retraining, prompt redesign, or online updating is introduced.

As illustrated in Fig.~\ref{fig:asvd_overview}, TSVD directly decomposes the pretrained projection weights as a closed-form baseline, whereas our ASVD first applies channel-wise scaling, decomposes the scaled weights, and then maps the factors back to the original operator space.
At a fixed rank, the two variants share the same inference-time structure and complexity; only factor generation differs.

\subsection{Unified Projection Representation}
\label{sec:unified_projection}

Transformer-based CIR models use heterogeneous projection operators.
A linear layer is parameterized by $W\in\mathbb{R}^{o\times i}$, where $i$ and $o$ denote the input and output channel dimensions.
For a convolution kernel $K\in\mathbb{R}^{o\times i\times k_h\times k_w}$, we use the matrix form
\begin{equation}
	W=\mathrm{reshape}(K)\in\mathbb{R}^{o\times (ik_hk_w)}.
	\label{eq:matricize}
\end{equation}
Eq.~\eqref{eq:matricize} lets linear layers and convolutional projections share the same decomposition procedure; after factorization, each operator is restored to its original layer type.

\subsection{Problem Formulation}
\label{sec:formulation}

Let $f(\cdot)$ denote a pretrained transformer-based blind CIR model.
Guided by the cost profile in Sec.~\ref{sec:projection_unit}, we select a set of projection operators $\mathcal{S}$ from the attention, feed-forward, convolutional projection, and prompt-conditioning pathways.
Each operator $m\in\mathcal{S}$ is represented by a weight matrix $W^{(m)}$.
Given a rank budget $r$, the goal is to construct
\begin{equation}
	W^{(m)} \approx \hat W^{(m)} = \hat W_a^{(m)}\hat W_b^{(m)},\qquad
	\mathrm{rank}(\hat W^{(m)}) \le r,
	\label{eq:lr_form}
\end{equation}
while leaving all non-selected modules unchanged.
The effective rank is bounded by the operator shape,
\begin{equation}
	r_{\mathrm{eff}}=\min(r,r_{\max}),
	\label{eq:reff}
\end{equation}
where $r_{\max}=\min(o,i)$ for a linear projection, and $r_{\max}=\min(o,ik_hk_w)$ for a matricized convolution. Thus, a shared $r$ is an upper bound rather than a common retained-rank fraction. It provides a controlled comparison of TSVD and ASVD without layer-wise rank optimization.

\subsection{Complexity of Low-rank Replacement}
\label{sec:complexity_form}

The weight and MAC counts follow directly from the matrix shape; the expressions below exclude unchanged bias terms.
For a dense projection $W\in\mathbb{R}^{o\times i}$, the original parameter count is $oi$.
Replacing it with two factors $W_a\in\mathbb{R}^{o\times r_{\mathrm{eff}}}$ and $W_b\in\mathbb{R}^{r_{\mathrm{eff}}\times i}$ changes the count to
\begin{equation}
	P_{\mathrm{lr}}=r_{\mathrm{eff}}(o+i).
	\label{eq:lr_params}
\end{equation}
The same expression gives the per-location multiply-accumulate cost of a linear projection.
For a convolution kernel reshaped as $W\in\mathbb{R}^{o\times (ik_hk_w)}$, the low-rank parameter count becomes
\begin{equation}
	P_{\mathrm{lr}}=r_{\mathrm{eff}}(o+ik_hk_w).
	\label{eq:conv_lr_params}
\end{equation}
Low-rank replacement reduces computation when the rank is small enough to offset the factorization overhead. The ablation study reports the resulting parameter and MAC counts.

\subsection{Implementation of Factorized Operators}
\label{sec:factorized_impl}

The low-rank factors are inserted as ordinary projection layers without intermediate nonlinearity or normalization.
For a linear projection, $W_b$ first maps the input feature from $i$ channels to $r_{\mathrm{eff}}$ channels, and $W_a$ then maps the intermediate feature from $r_{\mathrm{eff}}$ channels to $o$ channels.
If the original projection has a bias term, it is retained after the second factor to preserve the affine form.

For a convolutional projection, the reshaped factor $W_b\in\mathbb{R}^{r_{\mathrm{eff}}\times (ik_hk_w)}$ is implemented as a convolution from $i$ channels to $r_{\mathrm{eff}}$ channels with kernel size $k_h\times k_w$, and the second factor $W_a\in\mathbb{R}^{o\times r_{\mathrm{eff}}}$ as a $1\times 1$ convolution from $r_{\mathrm{eff}}$ channels to $o$ channels.
The original spatial stride, padding, and dilation are attached to the first factor, while the second factor mixes the intermediate channels.
This preserves the input-output tensor shape and leaves residual connections, attention reshaping, prompt fusion, and output heads unchanged.

Because TSVD and ASVD produce the same two-factor graph at the same rank, their difference is attributable to factor generation rather than runtime architecture.

\subsection{TSVD as the Training-free Low-rank Baseline}
\label{sec:tsvd}

For a selected weight matrix $W$, the singular value decomposition is
\begin{equation}
	W = U\Sigma V^\top,
	\label{eq:svd}
\end{equation}
where the singular values are sorted in descending order.
Here and below, the subscript $r$ denotes truncation to the effective rank $r_{\mathrm{eff}}$ in Eq.~\eqref{eq:reff}. TSVD retains these leading components:
\begin{equation}
	W \approx W_r = U_r\Sigma_r V_r^\top .
	\label{eq:svd_trunc}
\end{equation}
We form the two factors as
\begin{equation}
	W_a = U_r\Sigma_r,\qquad W_b = V_r^\top,
	\label{eq:merge}
\end{equation}
so that $W_r=W_aW_b$, and the original dense projection is replaced by two sequential low-rank projections.
This baseline is closed-form, deterministic, and training-free; however, its weight-centric objective ignores the channel magnitudes that compressed inputs induce.

\subsection{Proposed Activation-Calibrated SVD}
\label{sec:asvd}

\subsubsection{Activation-calibrated objective}
\label{sec:asvd_obj}

Our ASVD reshapes the decomposition objective using the channel-wise activation magnitudes observed in Sec.~\ref{sec:calibration_need}.
For a projection matrix $W$, we define diagonal scaling matrices
\begin{equation}
	D_l=\mathrm{diag}(d_l),\qquad D_r=\mathrm{diag}(d_r).
	\label{eq:diag}
\end{equation}
The scaling vectors $d_l$ and $d_r$ are derived from the output and input activation statistics, respectively, with dimensions matching the rows and columns of $W$.
For $E=W-\hat W$ and input $x$, the expected squared output perturbation is
\[
 \mathbb{E}\!\left[\|Ex\|_2^2\right]
 =\operatorname{tr}(E C_x E^\top),
\]
where $C_x=\mathbb{E}[xx^\top]$. Input activation magnitudes approximate input-channel importance, while output magnitudes weight errors in active output channels. Using these lightweight diagonal weights, ASVD minimizes
\begin{equation}
	\min_{\mathrm{rank}(\hat W)\le r}
	\left\|D_l (W-\hat W) D_r\right\|_F^2 .
	\label{eq:weighted_obj}
\end{equation}
Let $\tilde W=D_lWD_r$ and $Z=D_l\hat W D_r$. Invertible scaling preserves rank, so Eq.~\eqref{eq:weighted_obj} is equivalent to
\[
 \min_{\mathrm{rank}(Z)\le r}\|\tilde W-Z\|_F^2.
\]
Its truncated-SVD minimizer is
\begin{equation}
	\tilde W_r = \tilde U_r \tilde\Sigma_r \tilde V_r^\top ,
	\label{eq:tilde_svd}
\end{equation}
and mapping the factors back gives
\begin{equation}
	\begin{aligned}
 \hat W_a &= D_l^{-1}\tilde U_r\tilde\Sigma_r,\\
 \hat W_b &= \tilde V_r^\top D_r^{-1}.
 \end{aligned}
	\label{eq:weighted_merge}
\end{equation}
The product $\hat W=\hat W_a\hat W_b$ minimizes the weighted projection error in Eq.~\eqref{eq:weighted_obj} and replaces the original projection with two sequential factors. The scaling is absorbed during offline factor generation. We examine the contribution of each weighting side in Sec.~\ref{sec:ablation}.
After factor generation, ASVD executes in the same form as TSVD and adds no inference-time computation at a fixed rank.

\begin{algorithm}[t]
	\caption{Offline ASVD factor generation}
	\label{alg:asvd}
	\begin{algorithmic}[1]
		\REQUIRE Pretrained CIR model $f$, selected projection set $\mathcal{S}$, compressed calibration set $\mathcal{C}$, rank budget $r$
		\STATE Run $f$ on $\mathcal{C}$ and collect input and output channel magnitudes for each $m\in\mathcal{S}$
		\FOR{each selected projection $m\in\mathcal{S}$}
			\STATE Convert its weight to matrix form $W^{(m)}$
			\STATE Form diagonal scaling matrices $D_l^{(m)}$ and $D_r^{(m)}$ from the collected channel magnitudes
			\STATE Compute $\tilde W^{(m)}=D_l^{(m)}W^{(m)}D_r^{(m)}$
			\STATE Retain the largest $r_{\mathrm{eff}}$ singular values of $\tilde W^{(m)}$ and their singular vectors
			\STATE Map the factors back according to Eq.~\eqref{eq:weighted_merge}
			\STATE Replace the original dense projection with the two low-rank factors
		\ENDFOR
		\RETURN Compressed model with all other modules unchanged
	\end{algorithmic}
\end{algorithm}

In Algorithm~\ref{alg:asvd}, rank-$r_{\mathrm{eff}}$ SVD means the truncated product $\tilde U_{r_{\mathrm{eff}}}\tilde\Sigma_{r_{\mathrm{eff}}}\tilde V_{r_{\mathrm{eff}}}^{\top}$. For an $o\times i$ matrix, the two mapped factors have sizes $o\times r_{\mathrm{eff}}$ and $r_{\mathrm{eff}}\times i$, respectively.

\subsubsection{Offline calibration}
\label{sec:asvd_stats}

The scaling vectors $(d_l,d_r)$ are estimated by forwarding a small compressed calibration set through the pretrained model and accumulating per-channel mean absolute activation magnitudes at selected projection inputs and outputs.
For an input activation tensor $X^{(m)}$ and output activation tensor $Y^{(m)}$ of projection $m$, we compute the per-channel statistics
\begin{equation}
	\begin{aligned}
 a_{r,c}^{(m)}&=\mathbb{E}_{\mathcal{C},h,w}\!\left[|X_{c,h,w}^{(m)}|\right],\\
 a_{l,c}^{(m)}&=\mathbb{E}_{\mathcal{C},h,w}\!\left[|Y_{c,h,w}^{(m)}|\right].
 \end{aligned}
	\label{eq:activation_stats}
\end{equation}
where the expectation is taken over the calibration set $\mathcal{C}$ and the spatial dimensions.
For each side $s\in\{l,r\}$, we form strictly positive scaling entries as
\begin{equation}
 d_{s,c}^{(m)}=\max\!\left(a_{s,c}^{(m)},\eta\max_j a_{s,j}^{(m)}\right)+\epsilon,
 \label{eq:scale_protection}
\end{equation}
where $\eta=10^{-3}$ and $\epsilon=10^{-6}$. The relative floor limits inverse amplification on weak channels, and the additive term keeps the scaling invertible even when all channel statistics are zero. Statistics and factor generation use FP32. Channels with larger activation magnitudes retain larger weights in Eq.~\eqref{eq:weighted_obj}.
For convolutional projections, each input-channel statistic is repeated across the corresponding $k_hk_w$ entries of the matricized operator.
Calibration requires no ground-truth restored images, since it estimates feature magnitudes rather than a restoration loss.
By default, ASVD uses $16$ randomly sampled compressed images (all $14$ images for ICB), and its sensitivity to the calibration sample size is evaluated in Table~\ref{tab:sample_ablation}.

\subsection{Instantiation on PromptCIR}
\label{sec:instantiation_discussion}

We instantiate both variants on PromptCIR~\cite{li2024promptcir}, whose residual hybrid attention groups and transformer-style feed-forward blocks contain extensive projections across the encoder, bottleneck, decoder, and prompt-conditioning pathways.
Selected operators are replaced offline, while surrounding modules, prompts, normalization layers, nonlinearities, and inference procedures remain unchanged.
\section{Experiments}
\label{sec:experiments}

\subsection{Experimental Setup}
\label{sec:exp_setup}

We evaluate the framework on four blind compressed image restoration benchmarks: LIVE1~\cite{live1}, BSDS500~\cite{bsds}, ICB~\cite{ICB}, and DIV2K~\cite{DIV2K}.
Low-quality inputs are generated by JPEG compression with quality factors uniformly sampled from $[10,70]$, which spans mild to severe degradation.

We compare against JPEG, FBCNN~\cite{jiang2021fbcnn}, Contrastive JPEG~\cite{wang2022jpeg}, and prompt-based transformer frameworks including PromptIR~\cite{potlapalli2023promptir} and PromptCIR~\cite{li2024promptcir}.
All PromptCIR variants use identical ensemble inference and evaluation settings.

ASVD uses $16$ calibration images by default (all $14$ images for ICB). Calibration inputs are $128\times128$ JPEG images generated from the corresponding benchmark with QFs sampled uniformly from 10 to 70. Only compressed-input activations are used for factor generation; no retraining is required.

PSNR, SSIM, and PSNR-B are evaluated on RGB images and averaged over each dataset. PSNR uses the overall RGB mean squared error, whereas PSNR-B follows the channel-wise decibel averaging in SwinIR~\cite{liang2021swinir}.

Parameter counts and MACs are reported per model for a $256\times256$ input, including convolution, linear projections, and attention matrix multiplication.

\begin{table*}[t]
\caption{Quantitative comparison on compressed image restoration benchmarks. Results are PSNR / SSIM / PSNR-B ($\uparrow$). Best and second-best values at the displayed precision are bold and underlined. Parameter counts are per model; PromptCIR and its factorized variants use the same ensemble inference protocol.}
\label{tab:main_comparison}
\centering\small
\setlength{\tabcolsep}{4pt}
\begin{tabular}{@{}lcccccc@{}}
\toprule
\textbf{Method} & \textbf{Rank} & \textbf{Params. (M)} & \textbf{LIVE1} & \textbf{BSDS500} & \textbf{ICB} & \textbf{DIV2K}\\
\midrule
JPEG & -- & -- & 30.02/0.872/28.18 & 30.58/0.876/28.29 & 33.87/0.842/33.05 & 31.71/0.885/30.02 \\
FBCNN~\cite{jiang2021fbcnn} & -- & 71.92 & 32.08/0.904/31.44 & 32.41/0.903/31.45 & 35.90/0.865/36.33 & 33.96/0.918/33.41 \\
Contrastive JPEG~\cite{wang2022jpeg} & -- & 11.11 & 32.26/0.907/31.70 & \textbf{32.75}/\textbf{0.908}/\textbf{31.78} & 35.93/0.865/36.35 & 34.13/\underline{0.919}/33.62 \\
PromptIR~\cite{potlapalli2023promptir} & -- & 36.54 & 32.34/0.907/31.75 & \underline{32.67}/\underline{0.907}/\underline{31.67} & 36.35/\textbf{0.868}/36.75 & \textbf{34.38}/\textbf{0.922}/\textbf{33.81} \\
PromptCIR~\cite{li2024promptcir} & -- & 34.75 & \textbf{32.50}/\textbf{0.909}/\underline{31.91} & 32.23/0.906/31.35 & \textbf{36.49}/\underline{0.866}/\textbf{36.88} & \underline{34.25}/\underline{0.919}/\underline{33.71} \\
\midrule
\multirow{3}{*}{PromptCIR-TSVD} & 64 & 14.34 & 31.75/0.898/30.89 & 31.71/0.896/30.62 & 34.91/0.850/35.03 & 33.19/0.907/32.44 \\
 & 96 & 18.44 & 32.42/\underline{0.908}/\textbf{31.92} & 32.24/0.905/31.47 & 36.36/0.864/36.78 & 34.15/0.918/33.64 \\
 & 128 & 21.80 & 32.47/\textbf{0.909}/31.90 & 32.25/0.906/31.38 & 36.44/0.865/\underline{36.84} & 34.22/0.918/33.69 \\
\midrule
\multirow{3}{*}{PromptCIR-ASVD} & 64 & 14.34 & 31.67/0.899/31.04 & 31.74/0.898/30.84 & 34.96/0.853/35.27 & 33.10/0.907/32.56 \\
 & 96 & 18.44 & 32.43/\underline{0.908}/31.88 & 32.26/0.906/31.48 & 36.36/0.864/36.76 & 34.15/0.918/33.69 \\
 & 128 & 21.80 & \underline{32.48}/\textbf{0.909}/31.89 & 32.26/0.906/31.40 & \underline{36.45}/0.865/36.83 & 34.22/0.918/33.67 \\
\bottomrule
\end{tabular}
\end{table*}

\subsection{Comparison with Representative Methods}
\label{sec:comparison}

Table~\ref{tab:main_comparison} reports quantitative comparisons on the four benchmarks.
Both TSVD and ASVD substantially reduce the parameter count, with restoration fidelity increasing as the rank budget grows.

At rank~64, ASVD improves PSNR-B over TSVD by $0.12$--$0.24$ dB across all four benchmarks, with comparable PSNR and SSIM at identical parameter and MAC budgets. The PSNR differences between the two methods are below $0.10$ dB. At ranks~96 and~128, both methods approach the restoration performance of the original model.

ASVD reduces the parameter count by $59\%$ at rank~64. At rank~96, it retains a $47\%$ parameter reduction while keeping PSNR within $0.13$ dB of PromptCIR across all four benchmarks. On BSDS500, ranks~96 and~128 yield a small PSNR improvement of approximately $0.03$ dB. Such improvements are possible because minimizing projection error does not impose a fixed ordering on the final restoration error. The corresponding computational savings and measured latency are reported in Table~\ref{tab:complexity}.

\subsection{Qualitative and Feature-level Analysis}

\label{sec:qualitative_feature}
\begin{figure}[t]
	\centering
	\includegraphics[width=0.4\textwidth]{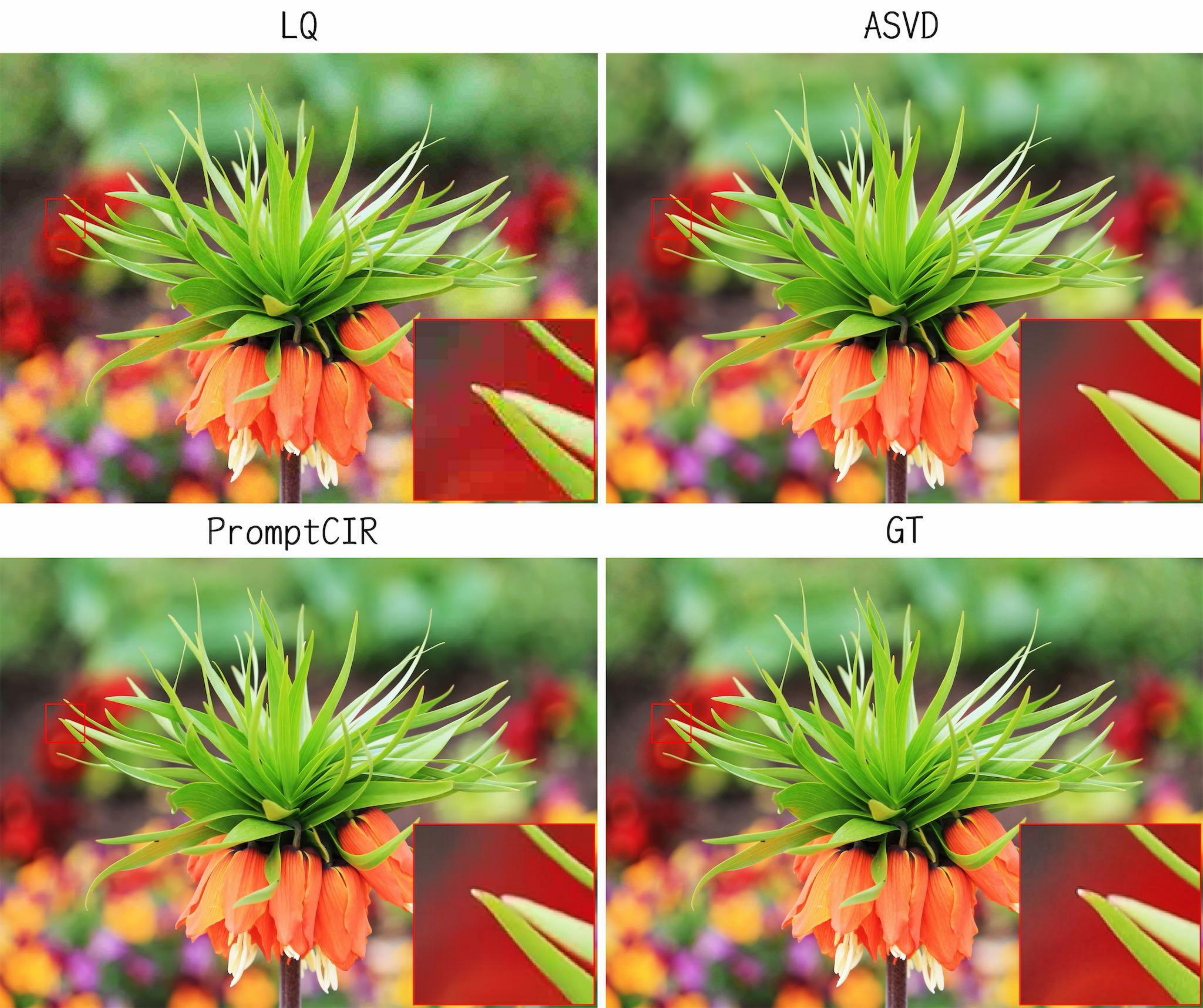}
	\caption{Visual comparison among the low-quality input (LQ), the ASVD output, the PromptCIR output, and the ground-truth reference (GT). Red boxes mark the zoomed region.}
	\label{fig:visual}
\end{figure}

Fig.~\ref{fig:visual} compares the low-quality input, the ASVD output, the PromptCIR output, and the ground-truth reference.
In the zoomed region, ASVD suppresses blocking artifacts while preserving object boundaries and local edge transitions.

\begin{figure}[htbp]
	\centering
	\subfloat[enc1: base vs TSVD]{
		\includegraphics[width=0.45\linewidth]{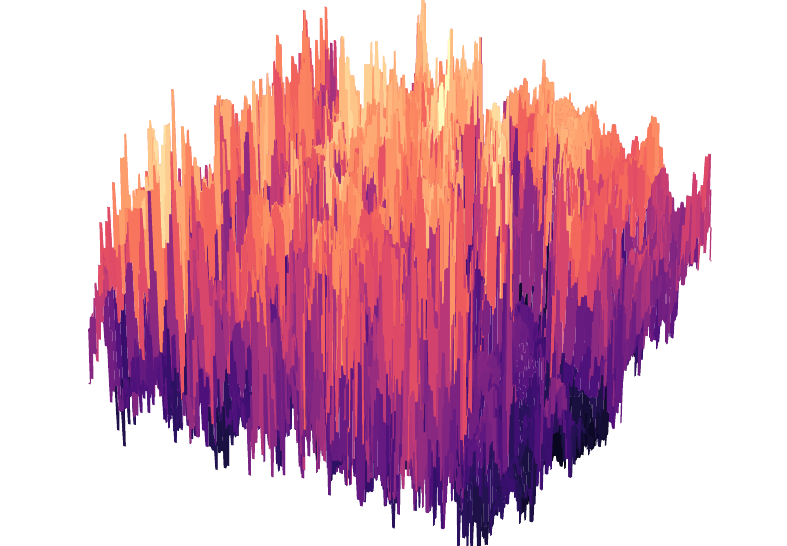}
	}
	\hfill
	\subfloat[enc1: base vs ASVD]{
		\includegraphics[width=0.45\linewidth]{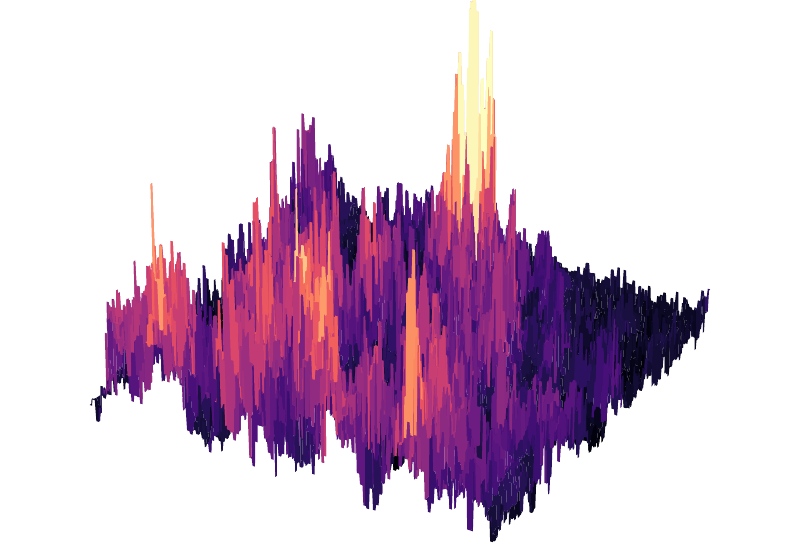}
	}
	
	\vspace{6pt}
	
	\subfloat[dec1: base vs TSVD]{
		\includegraphics[width=0.45\linewidth]{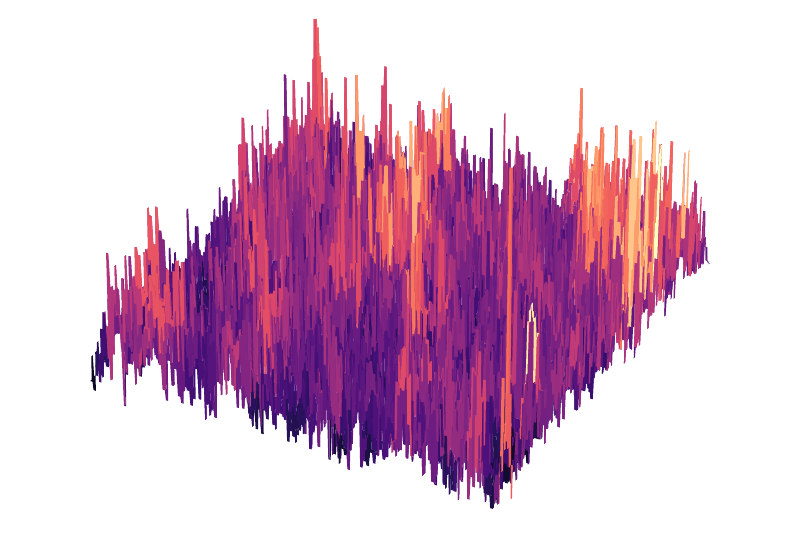}
	}
	\hfill
	\subfloat[dec1: base vs ASVD]{
		\includegraphics[width=0.45\linewidth]{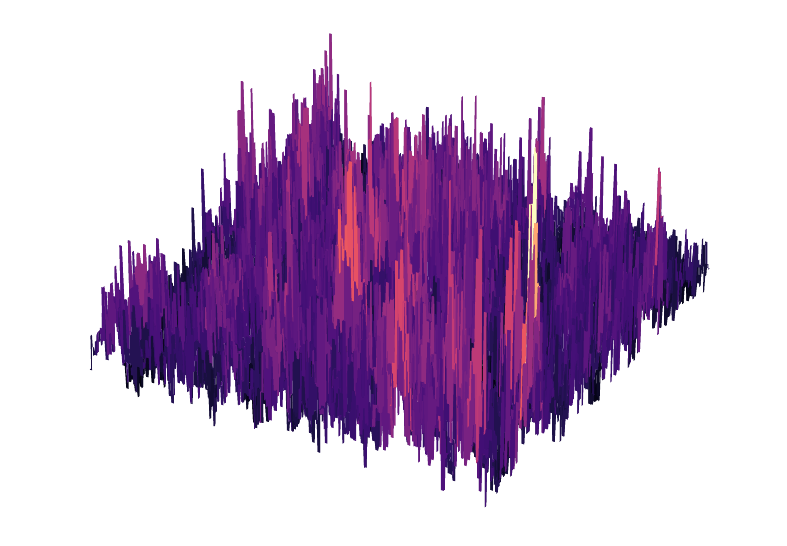}
	}
	
	\caption{Feature difference visualization on DIV2K at rank~$64$, using our ASVD instantiated on the PromptCIR backbone.
		Top row: encoder stage (\texttt{enc1}).
		Bottom row: decoder stage (\texttt{dec1}).
		Left: absolute feature error between the base model and TSVD.
		Right: absolute feature error between the base model and ASVD.
		Errors are computed on high-activation channels only.
		ASVD exhibits lower-magnitude and more spatially coherent error responses.}
	\label{fig:featdiff_enc_dec}
\end{figure}

Fig.~\ref{fig:featdiff_enc_dec} visualizes the absolute activation differences between the original PromptCIR backbone and the factorized models at encoder level~1 and decoder level~1.
Under rank~$64$, ASVD yields lower-magnitude and more spatially coherent feature errors on the high-activation channels, consistent with the proposed activation-weighted factor generation.

\Needspace{6\baselineskip}
\subsection{Ablation Study}
\label{sec:ablation}

\paragraph{Effect of rank}
Table~\ref{tab:rank_ablation} reports the rank ablation on DIV2K.
Increasing the rank generally improves restoration fidelity. ASVD achieves its largest PSNR gain over TSVD at rank~32 ($0.87$ dB). At ranks~16 and~64, it improves PSNR-B with PSNR differences below $0.12$ dB; the two methods approach one another at higher ranks.

\begin{table}[t]
\caption{Effect of rank on DIV2K. Results are PSNR / SSIM / PSNR-B ($\uparrow$).}
\label{tab:rank_ablation}
\centering\small
\setlength{\tabcolsep}{4pt}
\begin{tabular}{@{}ccc@{}}
\toprule
\textbf{Rank} & \textbf{TSVD} & \textbf{ASVD}\\
\midrule
8 & 28.26/0.854/28.21 & 28.34/0.846/28.39 \\
16 & 30.35/0.879/30.05 & 30.23/0.874/30.11 \\
32 & 30.82/0.889/30.61 & 31.69/0.892/31.36 \\
64 & 33.19/0.907/32.44 & 33.10/0.907/32.56 \\
96 & 34.15/0.918/33.64 & 34.15/0.918/33.69 \\
128 & 34.22/0.918/33.69 & 34.22/0.918/33.67 \\
\bottomrule
\end{tabular}
\end{table}

\Needspace{4\baselineskip}
\paragraph{Effect of calibration sample size}
\label{sec:sample_analysis}
Table~\ref{tab:sample_ablation} evaluates DIV2K at rank~64 over five random calibration samplings, with nested subsets and fixed test inputs. Results are reported as mean $\pm$ sample standard deviation. The PSNR standard deviation falls from $0.192$ dB with one image to $0.032$ dB with $16$ images. Performance approximately plateaus around $8$--$16$ images; increasing the sample count further does not produce a monotonic gain. These results support lightweight calibration and the default choice of $16$ images.

\begin{table}[t]
\caption{Calibration sample size on DIV2K (ASVD, rank~64). Mean $\pm$ sample standard deviation over five seeds.}
\label{tab:sample_ablation}
\centering\footnotesize
\setlength{\tabcolsep}{4pt}
\begin{tabular}{@{}cccc@{}}
\toprule
\textbf{Images} & \textbf{PSNR} & \textbf{SSIM} & \textbf{PSNR-B}\\
\midrule
1 & $32.861\!\pm\!0.192$ & $0.9053\!\pm\!0.0019$ & $32.314\!\pm\!0.157$ \\
2 & $33.006\!\pm\!0.069$ & $0.9058\!\pm\!0.0008$ & $32.429\!\pm\!0.090$ \\
4 & $33.064\!\pm\!0.059$ & $0.9065\!\pm\!0.0008$ & $32.495\!\pm\!0.044$ \\
8 & $33.073\!\pm\!0.032$ & $0.9067\!\pm\!0.0006$ & $32.518\!\pm\!0.026$ \\
16 & $33.062\!\pm\!0.032$ & $0.9070\!\pm\!0.0004$ & $32.518\!\pm\!0.028$ \\
32 & $33.041\!\pm\!0.043$ & $0.9070\!\pm\!0.0003$ & $32.511\!\pm\!0.034$ \\
\bottomrule
\end{tabular}
\end{table}

\Needspace{4\baselineskip}
\paragraph{Sensitivity to calibration quality range}
\label{sec:qf_sensitivity}
We vary the calibration QF range among $[10,30]$, $[40,70]$, and $[10,70]$, holding the sampled image identities fixed across the three ranges within each seed. Each model uses $16$ calibration images at rank~64 and is evaluated at fixed test QFs over three random samplings. Table~\ref{tab:qf_sensitivity} shows that the spread of mean PSNR across calibration ranges is at most $0.015$ dB on LIVE1 and $0.045$ dB on DIV2K for test QFs $10$--$70$. At the out-of-range test QF of $90$, the corresponding spreads remain $0.036$ and $0.053$ dB. The small changes indicate low sensitivity to the tested calibration quality distributions.

\begin{table}[t]
\caption{Sensitivity to the calibration QF range (rank~64, $16$ images). PSNR mean $\pm$ sample standard deviation over three seeds. QF~90 is outside all calibration ranges.}
\label{tab:qf_sensitivity}
\centering\footnotesize
\setlength{\tabcolsep}{4pt}
\begin{tabular}{@{}lcccc@{}}
\toprule
\textbf{Dataset} & \textbf{Test QF} & \textbf{$[10,30]$} & \textbf{$[40,70]$} & \textbf{$[10,70]$}\\
\midrule
\multirow{5}{*}{LIVE1} & 10 & $27.490\!\pm\!0.004$ & $27.487\!\pm\!0.006$ & $27.490\!\pm\!0.004$ \\
 & 30 & $31.280\!\pm\!0.008$ & $31.284\!\pm\!0.012$ & $31.284\!\pm\!0.013$ \\
 & 50 & $32.927\!\pm\!0.014$ & $32.935\!\pm\!0.012$ & $32.934\!\pm\!0.016$ \\
 & 70 & $34.447\!\pm\!0.031$ & $34.459\!\pm\!0.025$ & $34.453\!\pm\!0.035$ \\
 & 90 & $36.595\!\pm\!0.070$ & $36.630\!\pm\!0.051$ & $36.605\!\pm\!0.077$ \\
\midrule
\multirow{5}{*}{DIV2K} & 10 & $29.130\!\pm\!0.005$ & $29.128\!\pm\!0.005$ & $29.135\!\pm\!0.007$ \\
 & 30 & $32.908\!\pm\!0.021$ & $32.908\!\pm\!0.021$ & $32.922\!\pm\!0.019$ \\
 & 50 & $34.289\!\pm\!0.045$ & $34.297\!\pm\!0.042$ & $34.316\!\pm\!0.033$ \\
 & 70 & $35.558\!\pm\!0.061$ & $35.589\!\pm\!0.047$ & $35.602\!\pm\!0.049$ \\
 & 90 & $37.221\!\pm\!0.150$ & $37.249\!\pm\!0.100$ & $37.274\!\pm\!0.114$ \\
\bottomrule
\end{tabular}
\end{table}

\Needspace{4\baselineskip}
\paragraph{Contribution of the weighting sides}
\label{sec:weight_ablation}
Table~\ref{tab:weight_ablation} compares unweighted TSVD, input-only, output-only, and bilateral weighting on DIV2K at ranks~32 and~64. An input-only variant using the square root of the mean absolute magnitude is also included. All calibrated variants use $16$ images and three seeds. Common-reference output NMSE is evaluated on a common set of $16$ low-quality images, excluding the calibration identities and resized to $128\times128$.

At rank~32, bilateral weighting has the highest mean PSNR, SSIM, and PSNR-B among the tested variants. At rank~64, input-only weighting performs better, with the square-root variant attaining the highest restoration metrics and the lowest output NMSE. Thus, the output-side proxy is useful under the tighter rank budget but does not uniformly improve upon input-only scaling. The output NMSE and final restoration metrics need not induce the same ordering, because the former measures local projection errors rather than the nonlinear network output.

\begin{table*}[t]
\caption{Weighting-side ablation on DIV2K. Calibrated results are mean $\pm$ sample standard deviation over three seeds; TSVD is deterministic. Output NMSE ($\downarrow$) is measured on common reference inputs.}
\label{tab:weight_ablation}
\centering\small
\setlength{\tabcolsep}{4pt}
\begin{tabular}{@{}clcccc@{}}
\toprule
\textbf{Rank} & \textbf{Weighting} & \textbf{PSNR} & \textbf{SSIM} & \textbf{PSNR-B} & \textbf{Output NMSE}\\
\midrule
\multirow{5}{*}{32} & TSVD & $30.821\!\pm\!0.000$ & $0.8891\!\pm\!0.0000$ & $30.611\!\pm\!0.000$ & $0.0802\!\pm\!0.0000$ \\
 & Input only & $31.453\!\pm\!0.165$ & $0.8893\!\pm\!0.0004$ & $31.135\!\pm\!0.160$ & $0.0721\!\pm\!0.0003$ \\
 & Output only & $30.282\!\pm\!0.178$ & $0.8825\!\pm\!0.0002$ & $30.350\!\pm\!0.090$ & $0.0774\!\pm\!0.0001$ \\
 & Bilateral (ASVD) & $31.661\!\pm\!0.135$ & $0.8919\!\pm\!0.0013$ & $31.310\!\pm\!0.101$ & $0.0737\!\pm\!0.0004$ \\
 & Input only ($\alpha=0.5$) & $31.523\!\pm\!0.075$ & $0.8916\!\pm\!0.0009$ & $31.222\!\pm\!0.048$ & $0.0693\!\pm\!0.0001$ \\
\midrule
\multirow{5}{*}{64} & TSVD & $33.194\!\pm\!0.000$ & $0.9072\!\pm\!0.0000$ & $32.437\!\pm\!0.000$ & $0.0271\!\pm\!0.0000$ \\
 & Input only & $33.307\!\pm\!0.008$ & $0.9096\!\pm\!0.0004$ & $32.647\!\pm\!0.021$ & $0.0232\!\pm\!0.0001$ \\
 & Output only & $32.707\!\pm\!0.084$ & $0.9028\!\pm\!0.0002$ & $32.150\!\pm\!0.072$ & $0.0267\!\pm\!0.0000$ \\
 & Bilateral (ASVD) & $33.080\!\pm\!0.030$ & $0.9068\!\pm\!0.0005$ & $32.527\!\pm\!0.030$ & $0.0243\!\pm\!0.0001$ \\
 & Input only ($\alpha=0.5$) & $33.411\!\pm\!0.003$ & $0.9100\!\pm\!0.0000$ & $32.676\!\pm\!0.025$ & $0.0224\!\pm\!0.0000$ \\
\bottomrule
\end{tabular}
\end{table*}

\Needspace{4\baselineskip}
\paragraph{Uniform versus spectrum-based rank allocation}
\label{sec:rank_allocation}
A shared rank cap isolates the effect of calibrated factor generation, but does not imply equal compressibility of different operators. We therefore compare uniform rank~64 with non-uniform ranks selected by a common cumulative squared-singular-value energy threshold under a matched MAC budget. The search uses pretrained weights, not test restoration scores. Both allocations use the same DIV2K calibration set for all four benchmarks, so this controlled comparison differs from the dataset-specific calibration in Table~\ref{tab:main_comparison}.

Table~\ref{tab:allocation} shows PSNR gains of $0.008$--$0.409$ dB from spectrum-based allocation, with mixed SSIM and PSNR-B changes. The combined MAC budgets of the two models are $434.3195$G and $434.3178$G, respectively, but the parameter totals increase from $28.67$M to $42.37$M. Non-uniform allocation can therefore improve fidelity at matched arithmetic cost, while uniform ranks offer a smaller model. Neither allocation is claimed to be globally optimal.

\begin{table}[t]
\caption{Rank allocation at matched MACs. Results are PSNR / SSIM / PSNR-B ($\uparrow$). Both methods use the same fixed DIV2K calibration set.}
\label{tab:allocation}
\centering\small
\setlength{\tabcolsep}{4pt}
\begin{tabular}{@{}lcc@{}}
\toprule
\textbf{Dataset} & \textbf{Uniform 64} & \textbf{Spectrum-based}\\
\midrule
LIVE1 & 31.69/0.899/31.03 & 31.74/0.899/30.97 \\
BSDS500 & 31.71/0.898/30.81 & 31.72/0.898/30.69 \\
ICB & 34.79/0.847/35.12 & 35.20/0.853/35.40 \\
DIV2K & 33.10/0.907/32.56 & 33.28/0.908/32.61 \\
\bottomrule
\end{tabular}
\end{table}

\Needspace{4\baselineskip}
\paragraph{Sensitivity to the scaling floor}
\label{sec:floor_sensitivity}
On DIV2K at rank~64, varying the relative floor coefficient in Eq.~\eqref{eq:scale_protection} by $0.1\times$, $1\times$, and $10\times$ changes PSNR by at most $0.00016$ dB, SSIM by $0.000025$, and PSNR-B by $0.00267$ dB relative to the default. All resulting factors are finite. Restoration performance is therefore insensitive to these tested floor variations.

\Needspace{4\baselineskip}
\paragraph{Efficiency analysis}
\label{sec:efficiency}
Table~\ref{tab:complexity} reports model complexity and measured forward latency. At rank~64, the parameter count decreases from $34.75$M to $14.34$M and MACs from $258.79$G to $217.16$G. TSVD and ASVD have the same arithmetic cost at a fixed rank.

Forward latency is measured on an NVIDIA GeForce RTX~3090 in FP32 with batch size one and $256\times256$ inputs. We average three runs of $200$ synchronized forward passes after $50$ warm-up passes. Timing covers a single model without tiling or ensemble inference. Rank-64 ASVD reduces latency from $292.28$ to $238.42$ ms, yielding a $1.23\times$ speedup.

\begin{table}[t]
\caption{Complexity and $256\times256$ forward latency. Parameters and MACs are per model; MACs include attention matrix products. Latencies are in ms; speedup is ASVD relative to the original.}
\label{tab:complexity}
\centering\small
\setlength{\tabcolsep}{4pt}
\begin{tabular}{@{}cccccc@{}}
\toprule
\textbf{Rank} & \textbf{Params.} & \textbf{MACs} & \textbf{TSVD} & \textbf{ASVD} & \textbf{Speedup}\\ & (M) & (G) & (ms) & (ms) & ($\times$) \\
\midrule
32 & 9.79 & 164.15 & 208.26 & 213.19 & 1.37 \\
64 & 14.34 & 217.16 & 215.72 & 238.42 & 1.23 \\
96 & 18.44 & 254.38 & 228.53 & 240.56 & 1.21 \\
128 & 21.80 & 263.19 & 244.83 & 246.16 & 1.19 \\
\bottomrule
\end{tabular}
\end{table}
\section{Discussion}
\label{sec:discussion}

The empirical results support the analysis-driven acceleration scheme.
The profiling and spectral evidence in Fig.~\ref{fig:motivation_analysis} identify projection operators as the dominant source of cost and show that their weights contain exploitable low-rank structure.
Table~\ref{tab:main_comparison} demonstrates the effectiveness of activation-calibrated factor generation. At rank~64, ASVD improves PSNR-B over TSVD on all four benchmarks under identical parameter and MAC budgets.
This connects the motivation to evaluation: TSVD tests compact projection approximations, while the channel imbalance in Fig.~\ref{fig:motivation_analysis}(b) motivates activation-calibrated factor selection.

The rank ablation in Table~\ref{tab:rank_ablation} further clarifies the role of activation calibration.
At very small ranks, the factorized operators have limited capacity, and the choice of retained subspace determines most of the restoration error.
The weighted objective favors preservation of active channels. ASVD achieves its largest DIV2K PSNR gain over TSVD at rank~32 and consistently improves PSNR-B at rank~64, with comparable PSNR and SSIM.
As the rank grows, both decompositions recover a larger fraction of the pretrained operator, and the gap between TSVD and ASVD naturally narrows.
This trend distinguishes ASVD from runtime augmentation: it improves the offline factor selection under the same rank and complexity budget.
The feature visualizations in Fig.~\ref{fig:featdiff_enc_dec} provide representation-level evidence, with ASVD producing lower and more coherent errors on high-activation channels.

The calibration-size study in Table~\ref{tab:sample_ablation} supports the use of a small compressed calibration set.
The statistics used by our ASVD are per-channel activation magnitudes rather than task-loss gradients or reconstructed-image supervision; they simply measure which input and output channels respond more strongly when the pretrained model processes compressed images.
The multi-seed results indicate that a small set can stabilize these coarse channel weights, with performance approximately plateauing around $8$--$16$ images.
The quality-range study also shows low sensitivity to the tested calibration distributions. This stability supports blind CIR, where the model restores compressed inputs without receiving their JPEG quality factors.

The operator-level formulation in Fig.~\ref{fig:asvd_overview} keeps the accelerated model close to the original restoration backbone.
Each selected projection is replaced locally by two factors, while attention reshaping, prompt interaction, normalization, nonlinear activation, skip connections, and output reconstruction remain unchanged.
This locality avoids conflating low-rank replacement with architectural redesign and enables a controlled comparison: at a fixed rank, TSVD and ASVD share the same parameter count, MACs, tensor interfaces, and inference graph.
Consequently, their observed difference is attributable to activation-calibrated factor generation rather than topology or runtime computation.

The choice of the rank-$64$ operating point follows the same evidence in Tables~\ref{tab:rank_ablation} and~\ref{tab:complexity}.
It reduces parameters by $58.75\%$ and MACs by $16.09\%$ relative to the original PromptCIR backbone. Rank~96 reduces parameters by $46.93\%$ but MACs by only $1.70\%$; rank~128 exceeds the original MAC count by $1.70\%$.
Higher ranks are useful for demonstrating the convergence of TSVD and ASVD and for retaining restoration fidelity, but offer less arithmetic reduction.
Rank~64 provides a compact model with measurable PSNR-B improvements over TSVD, while rank~96 offers restoration performance close to the original backbone.
At rank~96, the PSNR difference from the original model ranges from $-0.13$ to $+0.03$ dB, illustrating the complementary higher-fidelity operating regime.

The method also offers a favorable deployment profile.
All factor generation is performed offline, and the compressed model retains the original PromptCIR topology, prompts, normalization layers, and restoration head.
ASVD is therefore complementary to pruning and quantization: it reduces the algebraic cost of dense projections while leaving precision reduction and sparsity-based acceleration available for deployment.
\section{Conclusion}
\label{sec:conclusion}

We presented a training-free framework for accelerating transformer-based blind compressed image restoration.
An empirical analysis of prompt-based restoration transformers shows that projection operators dominate the parameter and MAC cost, that their singular-value spectra exhibit operator-dependent low-rank structure, and that their compressed-image activations are strongly channel-imbalanced.
These observations support TSVD as a principled training-free baseline and explain why a uniform low-rank approximation can fall short under aggressive rank budgets.
Building on this rationale, we formulated projection-level replacement with TSVD as the closed-form baseline and designed ASVD as our activation-calibrated refinement that incorporates lightweight activation statistics into the decomposition objective.
The resulting factors are generated entirely offline from compressed calibration inputs while preserving the original restoration architecture and inference flow.
Experiments on four standard blind CIR benchmarks show that ASVD consistently improves PSNR-B over TSVD at rank~64 under identical parameter and MAC budgets, while maintaining comparable PSNR and SSIM.
These results indicate that activation-aware factor generation can compress restoration transformers without retraining data, task-specific supervision, or online adaptation.

The main evaluation uses a shared rank cap and one prompt-based backbone; a spectrum-based allocation study additionally examines non-uniform ranks under a matched MAC budget.
Promising future directions include further optimization of operator-wise or layer-wise rank allocation guided by spectral and activation profiles, combining ASVD with quantization and pruning for further compression, and extending it to other transformer-based restoration tasks.



\begin{thebibliography}{10}

\bibitem{ali2023vision}
Anas~M Ali, Bilel Benjdira, Anis Koubaa, Walid El-Shafai, Zahid Khan, and Wadii
  Boulila,
\newblock ``Vision transformers in image restoration: A survey,''
\newblock {\em Sensors}, vol. 23, no. 5, pp. 2385, 2023.

\bibitem{wang2022uformer}
Zhendong Wang, Xiaodong Cun, Jianmin Bao, Wengang Zhou, Jianzhuang Liu, and
  Houqiang Li,
\newblock ``Uformer: A general u-shaped transformer for image restoration,''
\newblock in {\em CVPR}, 2022, pp. 17683--17693.

\bibitem{zamir2022restormer}
Syed~Waqas Zamir, Aditya Arora, Salman Khan, Munawar Hayat, Fahad~Shahbaz Khan,
  and Ming-Hsuan Yang,
\newblock ``Restormer: Efficient transformer for high-resolution image
  restoration,''
\newblock in {\em CVPR}, 2022, pp. 5728--5739.

\bibitem{chen2023hat}
Xiangyu Chen, Xintao Wang, Wenlong Zhang, Xiangtao Kong, Yu~Qiao, Jiantao Zhou,
  and Chao Dong,
\newblock ``Hat: Hybrid attention transformer for image restoration,''
\newblock {\em arXiv preprint arXiv:2309.05239}, 2023.

\bibitem{li2024promptcir}
Bingchen Li, Xin Li, Yiting Lu, Ruoyu Feng, Mengxi Guo, Shijie Zhao, Li~Zhang,
  and Zhibo Chen,
\newblock ``Promptcir: Blind compressed image restoration with prompt
  learning,''
\newblock in {\em CVPR}, 2024, pp. 6442--6452.

\bibitem{eckart1936lora}
Carl Eckart and Gale Young,
\newblock ``The approximation of one matrix by another of lower rank,''
\newblock {\em Psychometrika}, vol. 1, no. 3, pp. 211--218, 1936.

\bibitem{liu2015data}
Xianming Liu, Xiaolin Wu, Jiantao Zhou, and Debin Zhao,
\newblock ``Data-driven sparsity-based restoration of jpeg-compressed images in
  dual transform-pixel domain,''
\newblock in {\em CVPR}, 2015, pp. 5171--5178.

\bibitem{dong2015arcnn}
Chao Dong, Yubin Deng, Chen~Change Loy, and Xiaoou Tang,
\newblock ``Compression artifacts reduction by a deep convolutional network,''
\newblock in {\em ICCV}, 2015, pp. 576--584.

\bibitem{wang2016d3}
Zhangyang Wang, Ding Liu, Shiyu Chang, Qing Ling, Yingzhen Yang, and Thomas~S
  Huang,
\newblock ``D3: Deep dual-domain based fast restoration of jpeg-compressed
  images,''
\newblock in {\em CVPR}, 2016, pp. 2764--2772.

\bibitem{zhang2018dmcnn}
Xiaoshuai Zhang, Wenhan Yang, Yueyu Hu, and Jiaying Liu,
\newblock ``Dmcnn: Dual-domain multi-scale convolutional neural network for
  compression artifacts removal,''
\newblock in {\em ICIP}. IEEE, 2018, pp. 390--394.

\bibitem{he2016deep}
Kaiming He, Xiangyu Zhang, Shaoqing Ren, and Jian Sun,
\newblock ``Deep residual learning for image recognition,''
\newblock in {\em CVPR}, 2016, pp. 770--778.

\bibitem{zhang2017dncnn}
Kai Zhang, Wangmeng Zuo, Yunjin Chen, Deyu Meng, and Lei Zhang,
\newblock ``Beyond a gaussian denoiser: Residual learning of deep cnn for image
  denoising,''
\newblock {\em IEEE transactions on image processing}, vol. 26, no. 7, pp.
  3142--3155, 2017.

\bibitem{mao2016red}
Xiaojiao Mao, Chunhua Shen, and Yu-Bin Yang,
\newblock ``Image restoration using very deep convolutional encoder-decoder
  networks with symmetric skip connections,''
\newblock {\em NeurIPS}, vol. 29, 2016.

\bibitem{tai2017memnet}
Ying Tai, Jian Yang, Xiaoming Liu, and Chunyan Xu,
\newblock ``Memnet: A persistent memory network for image restoration,''
\newblock in {\em ICCV}, 2017, pp. 4539--4547.

\bibitem{zhang2018rcan}
Yulun Zhang, Kunpeng Li, Kai Li, Lichen Wang, Bineng Zhong, and Yun Fu,
\newblock ``Image super-resolution using very deep residual channel attention
  networks,''
\newblock in {\em ECCV}, 2018, pp. 286--301.

\bibitem{zhang2019rnan}
Yulun Zhang, Kunpeng Li, Kai Li, Bineng Zhong, and Yun Fu,
\newblock ``Residual non-local attention networks for image restoration,''
\newblock {\em arXiv preprint arXiv:1903.10082}, 2019.

\bibitem{ehrlich2020qgac}
Max Ehrlich, Larry Davis, Ser-Nam Lim, and Abhinav Shrivastava,
\newblock ``Quantization guided jpeg artifact correction,''
\newblock in {\em ECCV}. Springer, 2020, pp. 293--309.

\bibitem{vaswani2017attention}
Ashish Vaswani, Noam Shazeer, Niki Parmar, Jakob Uszkoreit, Llion Jones,
  Aidan~N Gomez, {\L}ukasz Kaiser, and Illia Polosukhin,
\newblock ``Attention is all you need,''
\newblock {\em NIPS}, vol. 30, 2017.

\bibitem{liang2021swinir}
Jingyun Liang, Jiezhang Cao, Guolei Sun, Kai Zhang, Luc Van~Gool, and Radu
  Timofte,
\newblock ``Swinir: Image restoration using swin transformer,''
\newblock in {\em ICCV}, 2021, pp. 1833--1844.

\bibitem{hat}
Xiangyu Chen, Xintao Wang, Jiantao Zhou, Yu~Qiao, and Chao Dong,
\newblock ``Activating more pixels in image super-resolution transformer,''
\newblock in {\em CVPR}, 2023, pp. 22367--22377.

\bibitem{tsai2022stripformer}
Fu-Jen Tsai, Yan-Tsung Peng, Yen-Yu Lin, Chung-Chi Tsai, and Chia-Wen Lin,
\newblock ``Stripformer: Strip transformer for fast image deblurring,''
\newblock in {\em ECCV}. Springer, 2022, pp. 146--162.

\bibitem{chen2022simple}
Liangyu Chen, Xiaojie Chu, Xiangyu Zhang, and Jian Sun,
\newblock ``Simple baselines for image restoration,''
\newblock in {\em ECCV}. Springer, 2022, pp. 17--33.

\bibitem{guo2024mambair}
Hang Guo, Jinmin Li, Tao Dai, Zhihao Ouyang, Xudong Ren, and Shu-Tao Xia,
\newblock ``Mambair: A simple baseline for image restoration with state-space
  model,''
\newblock in {\em ECCV}. Springer, 2024, pp. 222--241.

\bibitem{li2023lsdir}
Yawei Li, Kai Zhang, Jingyun Liang, Jiezhang Cao, Ce~Liu, Rui Gong, Yulun
  Zhang, Hao Tang, Yun Liu, Denis Demandolx, et~al.,
\newblock ``Lsdir: A large scale dataset for image restoration,''
\newblock in {\em CVPR}, 2023, pp. 1775--1787.

\bibitem{DIV2K}
Eirikur Agustsson and Radu Timofte,
\newblock ``Ntire 2017 challenge on single image super-resolution: Dataset and
  study,''
\newblock in {\em CVPR workshops}, 2017, pp. 126--135.

\bibitem{ICB}
{Rawzor},
\newblock ``Image compression benchmark,'' [Online]. Available:
  \url{http://imagecompression.info/}.

\bibitem{yang2022aim}
Ren Yang, Radu Timofte, Xin Li, Qi~Zhang, Lin Zhang, Fanglong Liu, Dongliang
  He, Fu~Li, He~Zheng, Weihang Yuan, et~al.,
\newblock ``Aim 2022 challenge on super-resolution of compressed image and
  video: Dataset, methods and results,''
\newblock in {\em ECCV}. Springer, 2022, pp. 174--202.

\bibitem{jiang2021fbcnn}
Jiaxi Jiang, Kai Zhang, and Radu Timofte,
\newblock ``Towards flexible blind jpeg artifacts removal,''
\newblock in {\em ICCV}, 2021, pp. 4997--5006.

\bibitem{wang2022jpeg}
Xi~Wang, Xueyang Fu, Yurui Zhu, and Zheng-Jun Zha,
\newblock ``Jpeg artifacts removal via contrastive representation learning,''
\newblock in {\em ECCV}. Springer, 2022, pp. 615--631.

\bibitem{potlapalli2023promptir}
Vaishnav Potlapalli, Syed~Waqas Zamir, Salman~H Khan, and Fahad Shahbaz~Khan,
\newblock ``Promptir: Prompting for all-in-one image restoration,''
\newblock {\em NeurIPS}, vol. 36, pp. 71275--71293, 2023.

\bibitem{ma2023prores}
Jiaqi Ma, Tianheng Cheng, Guoli Wang, Qian Zhang, Xinggang Wang, and Lefei
  Zhang,
\newblock ``Prores: Exploring degradation-aware visual prompt for universal
  image restoration,''
\newblock {\em arXiv preprint arXiv:2306.13653}, 2023.

\bibitem{wang2024promptrr}
Tao Wang, Wanglong Lu, Kaihao Zhang, Wenhan Luo, Tae-Kyun Kim, Tong Lu,
  Hongdong Li, and Ming-Hsuan Yang,
\newblock ``Promptrr: Diffusion models as prompt generators for single image
  reflection removal,''
\newblock {\em arXiv preprint arXiv:2402.02374}, 2024.

\bibitem{pip}
Zilong Li, Yiming Lei, Chenglong Ma, Junping Zhang, and Hongming Shan,
\newblock ``Prompt-in-prompt learning for universal image restoration,''
\newblock {\em arXiv preprint arXiv:2312.05038}, 2023.

\bibitem{daclip}
Ziwei Luo, Fredrik~K Gustafsson, Zheng Zhao, Jens Sj{\"o}lund, and Thomas~B
  Sch{\"o}n,
\newblock ``Controlling vision-language models for universal image
  restoration,''
\newblock {\em arXiv preprint arXiv:2310.01018}, 2023.

\bibitem{visual_prompt_tuning}
Menglin Jia, Luming Tang, Bor-Chun Chen, Claire Cardie, Serge Belongie, Bharath
  Hariharan, and Ser-Nam Lim,
\newblock ``Visual prompt tuning,''
\newblock in {\em ECCV}. Springer, 2022, pp. 709--727.

\bibitem{prompt_learning}
Muhammad~Uzair Khattak, Hanoona Rasheed, Muhammad Maaz, Salman Khan, and
  Fahad~Shahbaz Khan,
\newblock ``Maple: Multi-modal prompt learning,''
\newblock in {\em CVPR}, 2023, pp. 19113--19122.

\bibitem{visual_prompt_tuning2}
Kihyuk Sohn, Huiwen Chang, Jos{\'e} Lezama, Luisa Polania, Han Zhang, Yuan Hao,
  Irfan Essa, and Lu~Jiang,
\newblock ``Visual prompt tuning for generative transfer learning,''
\newblock in {\em CVPR}, 2023, pp. 19840--19851.

\bibitem{li2024ucip}
Xin Li, Bingchen Li, Yeying Jin, Cuiling Lan, Hanxin Zhu, Yulin Ren, and Zhibo
  Chen,
\newblock ``Ucip: A universal framework for compressed image super-resolution
  using dynamic prompt,''
\newblock in {\em ECCV}. Springer, 2024, pp. 107--125.

\bibitem{ren2024mambacsr}
Yulin Ren, Xin Li, Mengxi Guo, Bingchen Li, Shijie Zhao, and Zhibo Chen,
\newblock ``Mambacsr: Dual-interleaved scanning for compressed image
  super-resolution with ssms,''
\newblock {\em arXiv preprint arXiv:2408.11758}, 2024.

\bibitem{guo2025compression}
Jinpei Guo, Zheng Chen, Wenbo Li, Yong Guo, and Yulun Zhang,
\newblock ``Compression-aware one-step diffusion model for jpeg artifact
  removal,''
\newblock in {\em ICCV}, 2025, pp. 14930--14939.

\bibitem{liang2021pruning}
Tailin Liang, John Glossner, Lei Wang, Shaobo Shi, and Xiaotong Zhang,
\newblock ``Pruning and quantization for deep neural network acceleration: A
  survey,''
\newblock {\em Neurocomputing}, vol. 461, pp. 370--403, 2021.

\bibitem{han2015learning}
Song Han, Jeff Pool, John Tran, and William Dally,
\newblock ``Learning both weights and connections for efficient neural
  network,''
\newblock {\em NeurIPS}, vol. 28, 2015.

\bibitem{han2015deep}
Song Han, Huizi Mao, and William~J Dally,
\newblock ``Deep compression: Compressing deep neural networks with pruning,
  trained quantization and huffman coding,''
\newblock {\em arXiv preprint arXiv:1510.00149}, 2015.

\bibitem{liu2017learning}
Zhuang Liu, Jianguo Li, Zhiqiang Shen, Gao Huang, Shoumeng Yan, and Changshui
  Zhang,
\newblock ``Learning efficient convolutional networks through network
  slimming,''
\newblock in {\em ICCV}, 2017, pp. 2736--2744.

\bibitem{denton2014exploiting}
Emily~L Denton, Wojciech Zaremba, Joan Bruna, Yann LeCun, and Rob Fergus,
\newblock ``Exploiting linear structure within convolutional networks for
  efficient evaluation,''
\newblock {\em NeurIPS}, vol. 27, 2014.

\bibitem{jaderberg2014speeding}
Max Jaderberg, Andrea Vedaldi, and Andrew Zisserman,
\newblock ``Speeding up convolutional neural networks with low rank
  expansions,''
\newblock {\em arXiv preprint arXiv:1405.3866}, 2014.

\bibitem{lebedev2014speeding}
Vadim Lebedev, Yaroslav Ganin, Maksim Rakhuba, Ivan Oseledets, and Victor
  Lempitsky,
\newblock ``Speeding-up convolutional neural networks using fine-tuned
  cp-decomposition,''
\newblock {\em arXiv preprint arXiv:1412.6553}, 2014.

\bibitem{wang2024svd}
Xin Wang, Yu~Zheng, Zhongwei Wan, and Mi~Zhang,
\newblock ``Svd-llm: Truncation-aware singular value decomposition for large
  language model compression,''
\newblock {\em arXiv preprint arXiv:2403.07378}, 2024.

\bibitem{chekalina2025fwsvd}
Viktoriia Chekalina, Daniil Moskovskiy, Daria Cherniuk, Maxim Kurkin, Andrey
  Kuznetsov, and Evgeny Frolov,
\newblock ``Generalized fisher-weighted svd: Scalable kronecker-factored fisher
  approximation for compressing large language models,''
\newblock {\em arXiv preprint arXiv:2505.17974}, 2025.

\bibitem{ma2024deepcache}
Xinyin Ma, Gongfan Fang, and Xinchao Wang,
\newblock ``Deepcache: Accelerating diffusion models for free,''
\newblock in {\em CVPR}, 2024, pp. 15762--15772.

\bibitem{wimbauer2024blockcache}
Felix Wimbauer, Bichen Wu, Edgar Schoenfeld, Xiaoliang Dai, Ji~Hou, Zijian He,
  Artsiom Sanakoyeu, Peizhao Zhang, Sam Tsai, Jonas Kohler, et~al.,
\newblock ``Cache me if you can: Accelerating diffusion models through block
  caching,''
\newblock in {\em CVPR}, 2024, pp. 6211--6220.

\bibitem{chen2025icc}
Zhiyuan Chen, Keyi Li, Yifan Jia, Le~Ye, and Yufei Ma,
\newblock ``Accelerating diffusion transformer via increment-calibrated caching
  with channel-aware singular value decomposition,''
\newblock in {\em CVPR}, 2025, pp. 18011--18020.

\bibitem{selvaraju2024fora}
Pratheba Selvaraju, Tianyu Ding, Tianyi Chen, Ilya Zharkov, and Luming Liang,
\newblock ``Fora: Fast-forward caching in diffusion transformer acceleration,''
\newblock {\em arXiv preprint arXiv:2407.01425}, 2024.

\bibitem{ma2024learning}
Xinyin Ma, Gongfan Fang, Michael Bi~Mi, and Xinchao Wang,
\newblock ``Learning-to-cache: Accelerating diffusion transformer via layer
  caching,''
\newblock {\em NeurIPS}, vol. 37, pp. 133282--133304, 2024.

\bibitem{huang2024harmonica}
Yushi Huang, Zining Wang, Ruihao Gong, Jing Liu, Xinjie Zhang, Jinyang Guo,
  Xianglong Liu, and Jun Zhang,
\newblock ``Harmonica: Harmonizing training and inference for better feature
  caching in diffusion transformer acceleration,''
\newblock {\em arXiv preprint arXiv:2410.01723}, 2024.

\bibitem{chen2024delta}
Pengtao Chen, Mingzhu Shen, Peng Ye, Jianjian Cao, Chongjun Tu, Christos-Savvas
  Bouganis, Yiren Zhao, and Tao Chen,
\newblock ``{$\Delta$-DiT}: A training-free acceleration method tailored for
  diffusion transformers,''
\newblock {\em arXiv preprint arXiv:2406.01125}, 2024.

\bibitem{zou2024accelerating}
Chang Zou, Xuyang Liu, Ting Liu, Siteng Huang, and Linfeng Zhang,
\newblock ``Accelerating diffusion transformers with token-wise feature
  caching,''
\newblock {\em arXiv preprint arXiv:2410.05317}, 2024.

\bibitem{live1}
H.~R. Sheikh, Z.~Wang, L.~Cormack, and A.~C. Bovik,
\newblock ``{LIVE} image quality assessment database release 2,'' [Online].
  Available: \url{https://live.ece.utexas.edu/research/quality/}, 2005.

\bibitem{bsds}
David Martin, Charless Fowlkes, Doron Tal, and Jitendra Malik,
\newblock ``A database of human segmented natural images and its application to
  evaluating segmentation algorithms and measuring ecological statistics,''
\newblock in {\em ICCV}. IEEE, 2001, vol.~2, pp. 416--423.

\end{thebibliography}
\end{document}